\documentclass[10pt,twocolumn,letterpaper]{article}

\usepackage{cvpr}              

\usepackage{graphicx}
\usepackage{amsmath}
\usepackage{rotating}
\usepackage{amssymb}
\usepackage{booktabs}
\usepackage{algorithm2e}
\usepackage{placeins}
\SetKwComment{Comment}{/* }{ */}
\RestyleAlgo{ruled} 
\usepackage{float}
\restylefloat{table}
\usepackage[pagebackref,breaklinks,colorlinks]{hyperref}

\usepackage[capitalize]{cleveref}
\crefname{section}{Sec.}{Secs.}
\Crefname{section}{Section}{Sections}
\Crefname{table}{Table}{Tables}
\crefname{table}{Tab.}{Tabs.}

\def\cvprPaperID{26} 
\def\confName{CVPR}
\def\confYear{2022}

\begin{document}

\title{Proposal-free Lidar Panoptic Segmentation with Pillar-level Affinity}

\author{Qi Chen\thanks{work done while interning at Motional}\\
The Johns Hopkins University\\
{\tt\small qchen42@jhu.edu}
\and
Sourabh Vora\\
Motional\\
{\tt\small sourabh.vora@motional.com}
}
\maketitle

\begin{abstract}

    We propose a simple yet effective proposal-free architecture for lidar panoptic segmentation. We jointly optimize both semantic segmentation and class-agnostic instance classification in a single network using a pillar-based bird’s-eye view representation. The instance classification head learns pairwise affinity between pillars to determine whether the pillars belong to the same instance or not. We further propose a local clustering algorithm to propagate instance ids by merging semantic segmentation and affinity predictions. Our experiments on nuScenes dataset show that our approach outperforms previous proposal-free methods and is comparable to proposal-based methods which requires extra  annotation from object detection.
    
\end{abstract}

\section{Introduction}
\label{sec:intro}
Panoptic Segmentation \cite{Kirillov_2019_CVPR} unifies the two tasks of semantic segmentation (assign a class label to each pixel) and instance segmentation (detect and segment each object instance). In a similar vein, pointcloud panoptic segmentation needs to provide the semantic label id for each point and further assign a unique instance label to points that belong to the same object.

Inspired by image panoptic segmentation literature, lidar panoptic segmentation algorithms can be categorized into two categories: proposal-based and proposal-free. While proposal-based methods typically require complicated and slower object detection heads, proposal-free methods usually adopt a bottom-up approach and group semantic segmentation outputs efficiently.

EfficientLPS\cite{sirohi2021efficientlps} and PolarStream\cite{chen2021polarstream} generate proposals and group points using the bounding box centers as cluster centers. These proposal-based methods have three issues: 1) the post-processing module needs to resolve conflicts between different branches, such as the inconsistent class labels of semantic branch and instance branch prediction; 2) the training and inference is not efficient because object detection/instance segmentation and semantic segmentation are coupled tasks and have redundant information; 3) overlapping boxes result in over-segmentation. 

\begin{figure}
\centering
\includegraphics[width=0.4\textwidth]{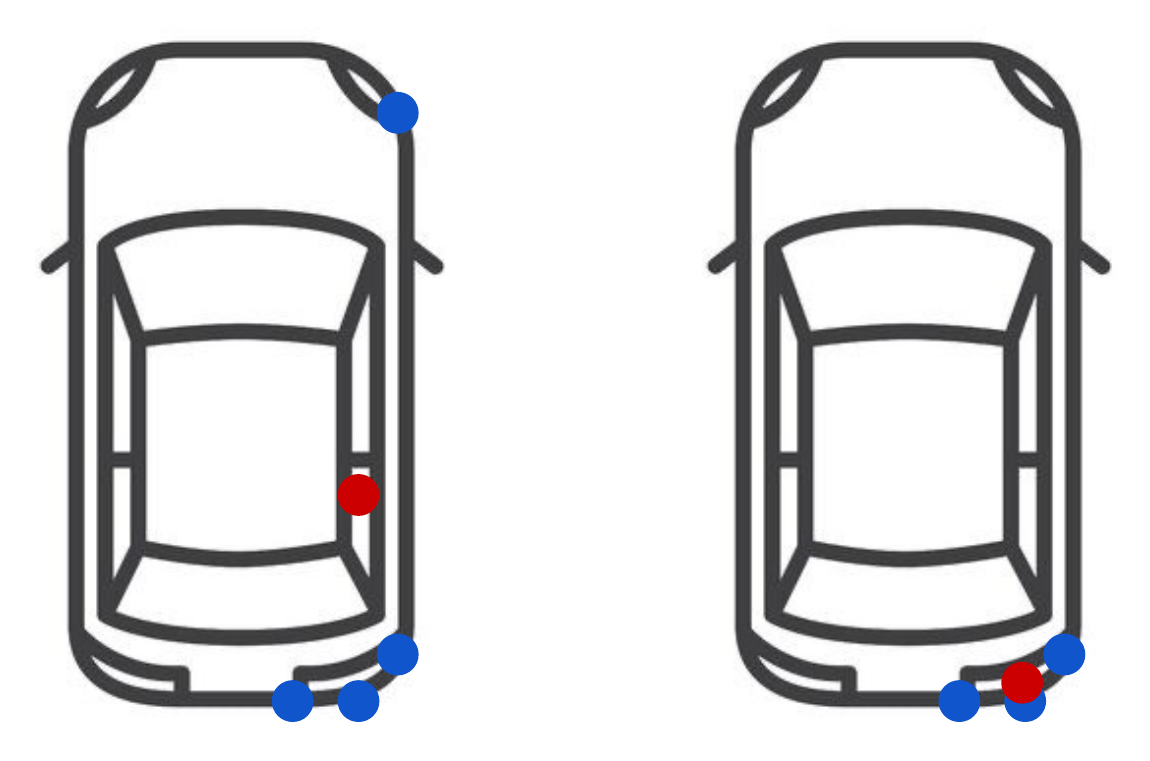} 
  \caption{The issue of centroid-based proposal-free lidar panoptic segmentation approach. Miss of a single observation (blue) by lidar scan can cause significant change in centroid (red).}
  \label{fig:center-based}
\end{figure}

To avoid these conflicts, proposal-free methods start with semantic segmentation predictions and then generate instance masks through grouping or clustering. Panoptic-PolarNet\cite{fong2021panoptic} uses a class-agnostic instance segmentation branch to predict the centroid of points coming from same instances and regresses the offset from each point in those instances to their centroid. The predicted class-agnostic centroids serve as cluster centers. This centroid-based approach is challenging, particularly for pointclouds of partially occluded objects. As shown in Fig. \ref{fig:center-based}, even a miss of a point may result in a dramatic change in the centroid. Furthermore, since the predicted centroids are shared and reused by multiple semantic classes, there is a risk of over-segmentation. And finally the offset estimation sub-task has its own limitations. It is harder to learn because of its large variance and is not synergistic to the semantic segmentation task, thereby competing against it in gradient distribution.

\begin{figure*}
  \centering
  \begin{subfigure}{0.45\textwidth}
    \includegraphics[scale=0.7]{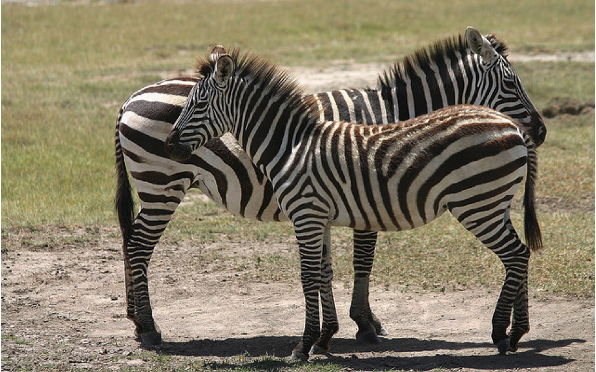} 
    \caption{Original image from COCO dataset}
    \label{fig:short-a}
  \end{subfigure}
  \begin{subfigure}{0.45\textwidth}
    \includegraphics[scale=0.36]{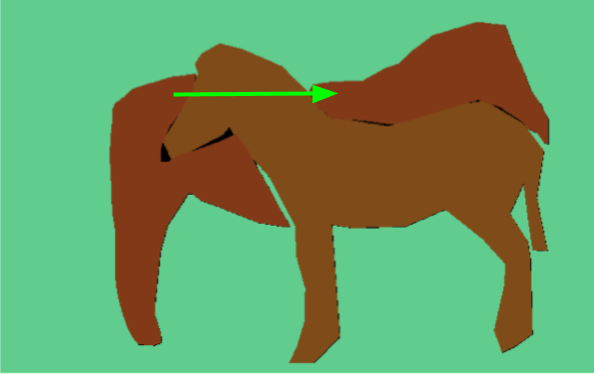} 
    \caption{Panoptic labels for (a)}
    \label{fig:short-b}
  \end{subfigure}
  \begin{subfigure}{0.45\textwidth}
    \includegraphics[]{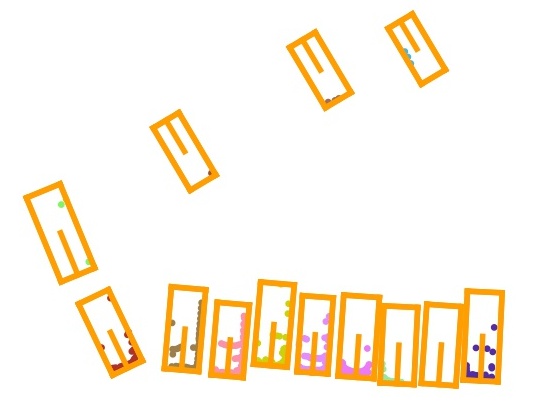} 
    \caption{Lidar points (on BEV) and bounding box annotation for cars from NuScenes dataset}
    \label{fig:short-c}
  \end{subfigure}
  \begin{subfigure}{0.45\textwidth}
    \includegraphics[scale=0.42]{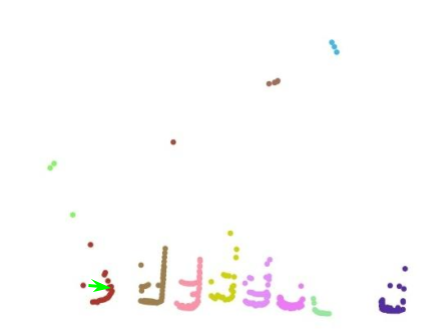} 
    \caption{Panoptic labels for (c).}
    \label{fig:short-d}
  \end{subfigure}
  \caption{Long range dependency vs short range dependency for thing classes. The green arrows show long range dependency on an image and short range dependency on a lidar based bird's-eye view representation.}
  \label{fig:short}
\end{figure*}

In contrast to unreliable instance centroids, the affinity describing pairwise semantic relations between pixels is invariant to the change of object appearances. The affinity matrix is often used in semantic segmentation to encourage the labeling consistency of local similar pixels. This motivates us to learn affinities for the instance segmentation task as it will help group pillars from the same instance and discriminate between different instances. However, the dimensionality of the resulting affinity matrix will be very high. Pairwise similarity between all pixels, i.e. pillars in the bird's-eye view grid, requires a holistic affinity matrix of size $(H \times W) \times (H \times W)$ for the grid with spatial size $H \times W$. On images, a holistic affinity matrix is needed due to long range dependency between pixels. As shown in Fig. \ref{fig:short-b}, objects in images have varying scales and can be occluded by another instance of the same semantic class so pixels far apart may share the same instance id. This is however not a problem with the lidar bird’s-eye view representation. As shown in Fig. \ref{fig:short-d} objects appear in fixed scales and objects with same semantic class usually do not overlap with each other. This means that only the pillar-level affinity in a local neighborhood is needed to group semantic predictions of the same class. We further observe that if we scan the bird's-eye view grid from left to right, as the green arrow in Fig. \ref{fig:short-d}, a pillar belonging to a thing class can either be the start of an instance or share the same instance id as the pillars to its left. Based on this observation, we narrow our focus on the affinity between a pillar $pl_i$ and another pillar to the left of $pl_i$. This allows us to reduce the affinity associated with $pl_i$ from a vector of length $HW$ to a single value, and therefore reduce the affinity matrix of size $(H \times W) \times (H \times W)$ into a matrix of size $H \times W$.

In this paper, we propose a panoptic segmentation model consisting of two heads: a semantic segmentation head and a head which predicts pillar-level affinities. The affinity head performs pillar-wise binary classification to predict whether the pillar belongs to the same instance as one of its previous pillars. We also propose a zig-zag traversal order for all pillars. The learned affinities and the zig-zag traversal order enable us to propose a local clustering algorithm to assign unique instance ids to pillars with zero affinity predictions and propagate the instance ids to the similar pillars.

To summarize, our contributions are three-fold:
\begin{itemize}
  \item We simplify lidar panoptic segmentation training into purely classification sub-tasks: a) semantic segmentation and b) pillar-wise affinity prediction where we reduce the dimensionality of the affinity matrix
  \item We propose a local clustering algorithm for instance id propagation
  \item Our approach closes the gap between proposal-based and proposal-free methods and outperforms previous proposal-free methods
\end{itemize}


\begin{figure*}
  \centering
\includegraphics[width=\textwidth]{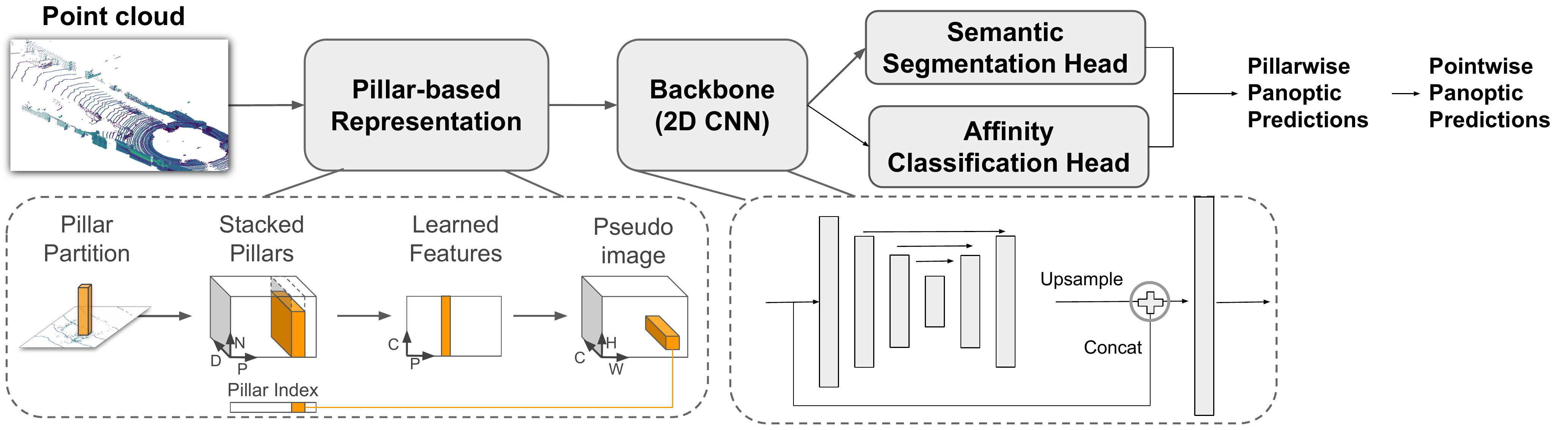} 
  \caption{Panoptic network overview. Figure inspired by PointPillars\cite{lang2019pointpillars}.}
  \label{fig:net}
\end{figure*}

\section{Related Works}
\subsection{Image-based Panoptic Segmentation}
Since panoptic segmentation is a unified task of semantic segmentation and instance segmentation, most panoptic segmentation models adopt a detection followed by semantic segmentation approach, or semantic segmentation followed by grouping approach. Thus panoptic segmentation methods can be generally categorized as top-down (proposal-based) methods and bottom-up (proposal-free) methods. Most proposal-based methods, such as Panoptic FPN\cite{Kirillov_2019_CVPR}, start from instance segmentation, usually with a Mask R-CNN\cite{he2017mask} and FPN\cite{lin2017feature}. The two branches are coupled with redundant information. The post-processing module needs to merge the output results of the semantic branch and the instance branch. Sometimes it needs to resolve conflicts between different branches such as the inconsistent class labels. UPSNet\cite{xiong2019upsnet} developed a parameter-free panoptic head for unified training and inference. A line of transformer-based works\cite{wang2020axial,hu2021istr} directly generate masks as set prediction, eliminating the need for merging operators. On the other hand, Axial-DeepLab\cite{wang2020axial}, the state-of-the-art bottom-up approach that predicts pixel-wise offsets to pre-defined instance centers, suffers from highly deformable objects or near-by objects with close centers. 



\subsection{Lidar Panoptic Segmentation}
Similar to image-based approaches, lidar panoptic segmentation methods, though less explored, can also be categorized into proposal-based and proposal-free methods. Adapted from a 2D panoptic segmentation architecture, EfficientLPS\cite{sirohi2021efficientlps} generates region proposals from encoded features and then detects the instances in parallel to perform semantic segmentation. PolarStream\cite{chen2021polarstream} does semantic segmentation and object detection simultaneously and then groups points according to bounding box centers. Proposal-free methods generally infer semantic segmentation before detecting instances through either keypoint detection tasks like center point estimation or through voting/learnable clustering in the range view or bird's-eye view domain. In addition to semantic segmentation, Panoptic-PolarNet\cite{fong2021panoptic} trains an instance head that predicts the centroid of the points in each instance and offsets from each point to the centroid. The proposal-free architecture is trained without any bounding box annotation.

\subsection{Affinity Learning for Semantic Segmentation}
Several works adopt graphical models with affinity as binary potentials to refine the outputs of fully supervised semantic segmentation models. RWN\cite{bertasius2017convolutional} learns pixelwise affinity and semantic segmentation jointly and employs a convolutional random walk layer to combine both objectives. LS-DeconvNet\cite{cheng2017locality} leverages affinity to refine object boundaries in upsampling layers. Another line of works incorporates affinity in a weakly supervised setting. Given only image-level class annotation, Ahn \etal\cite{ahn2018learning} predicts affinities by a deep neural network and propagates labels to nearby areas which belong to the same semantic entity.

\section{Problem Statement}
A lidar point cloud sample consists of a set of N points, each point represented by a vector of point feature $f =(x,y,z,i,t)$, where $(x,y,z)$ is its location in Cartesian coordinates relative to the sensor, $i$ is the reflection intensity and $t$ is the timestamp when the LiDAR point is captured. LiDAR point cloud semantic segmentation task aims to predict a set of class labels $S=\{s_i\in \{1,2,...,K\}\}_{i=1,...,N}$ for the points. Panoptic segmentation task extends this problem to requiring different instance ids assigned to points belonging to countable instances in some `thing' classes, e.g., car, bicycle, and pedestrian. The remaining classes are `stuff' classes, which do not require detailed separation and share the same label among all points.
\begin{figure*}
  \centering
  \begin{subfigure}{0.31\linewidth}
    \includegraphics[width=0.9\linewidth]{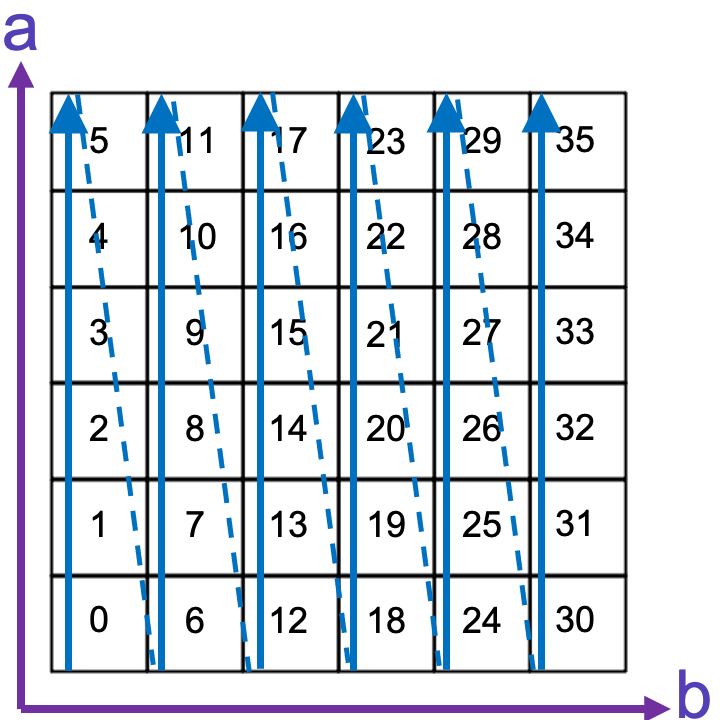}
   \caption{Zigzag traversal order for two-dimensional pillars (along a and b axes) on BEV. }
    \label{fig:zigzag}
  \end{subfigure}
  \hfill
  \begin{subfigure}{0.67\linewidth}
    \includegraphics[width=\linewidth]{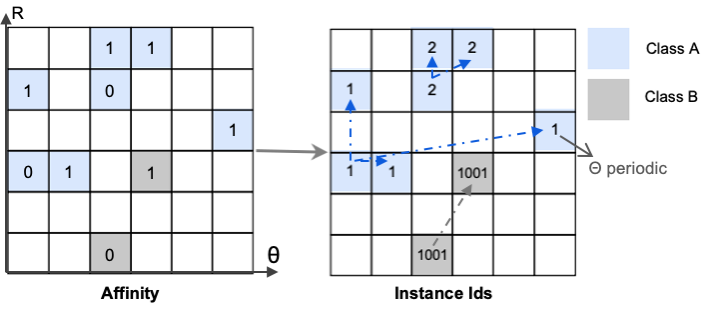}
   \caption{Illustration of instance id propagation from semantic and affinity prediction. Here shows an example of polar pillars with size $6 \times 6$. Wrap-around effect is considered because $\theta$ is periodic. }
   \label{fig:clustering}
  \end{subfigure}
  \caption{Pillar-level affinity Representation.}
\end{figure*}
\paragraph{Zig-zag Traversal Order for Pillars} To facilitate our representation of affinity, we first define an order for the two-dimensional pillars on BEV, as show in  Fig. \ref{fig:zigzag}. We uniquely map pillars along $a,b$ axes to one-dimensional order. For Cartesian pillars, $a,b$ axes corresponds to $y,x$ axes respectively ($x, y$ axes are also feasible); for Polar pillars, $a, b$ axes corresponds to r, $\theta$ axes respectively. 

\section{Approach}

As shown in Fig. \ref{fig:net}, our approach consists of the following four components: (1) a small pillar feature encoder that encodes the raw point cloud data to 2D pillar-based representation; (2) a shared 2D CNN backbone; (3) two heads for semantic segmentation and affinity classification (4) a local clustering scheme for instance id propagation.

\subsection{Preliminary}
\subsubsection{Pillar-Based Point Clouds Representation}
There are two alternative pillar partition schemes, cartesian pillars and polar pillars, based on which coordinate system is chosen. The cartesian pillars are rectangular-shaped on the bird's-eye view while the polar pillars are wedge-shaped. In this paper, we present the results of both pillar partition schemes.

\paragraph{Cartesian Pillar Partition}
The point cloud is discretized into an evenly spaced grid in the $x-y$ plane, creating a set of pillars. Each pillar contains T points where T varies for different pillars. A Pillar Feature Encoder consisting of a multi-layer perceptron and a max pooling layer is applied to T point features in each pillar to extract a single vector of dimensions $C$ as the pillar feature. The encoded pillars form a bird's-eye view pseudo-image of size $H \times W \times C$.

\paragraph{Polar Pillar Partition}
Each point is represented by a vector of point feature $f = (r, \theta, z, x, y, i, t)$, where $(r, \theta)$ is the polar coordinates. The points are grouped in $r-\delta$ plane, resulting in a BEV pseudo-image of  size $R \times \Theta \times C$.
\subsubsection{2D CNN Backbone}
We use a similar backbone as \cite{chen2021polarstream} and the structure is shown in Fig.\ref{fig:net}. The backbone has a UNet-like \cite{ronneberger2015u} structure, with a top-down network that produces features at increasingly small spatial resolution and a network that performs deconvolution and concatenation of the top-down features. The final output features are a concatenation of all upsampled features that originated from different strides as well as the input pseudo-image.
\subsubsection{Pillarwise Semantic Segmentation}
 The semantic segmentation head is doing pillarwise K-class classification, with output size $H \times W \times K$, same spatial size as the input pillars. The ground truth label for each pillar is decided by majority voting of points within the same pillar. During inference, all points inside the same pillar share the same predicted semantic class.

\subsection{Pillar-level Affinity}

\paragraph{Class-agonistic Pillar-level Affinity} 

We denote affinity between two pillars $pl_i$ and $pl_j$ as $a_{pl_i, pl_j}$. The holistic affinity vector associated with $pl_i$ is 
\begin{equation}
    A_i = \{a_{pl_i, pl_j}\}, j=0,1,...,H \times W-1
\end{equation} As discussed in Sec. \ref{sec:intro}, we only care about whether $pl_i$ shares the same instance as any of its previous pillars in our zig-zag traversal order. Our target for $pl_i$ can be reduced as
\begin{equation}
    a_i' = max(\{a_{pl_i, pl_j}\}), j=0,1,...,i-1
\end{equation}
which is a single value. The ground truth for $a_i'$ is a binary value. $a_i'=0$ means $pl_i$ is not similar to any of its previous pillars and $a_i'=1$ means $pl_i$ shares the same instance with at least one of its previous pillars. We can also see that if $a_i'=0$, $pl_i$ is the start (left-most pillar) of a new instance. 

\paragraph{Generating Affinity Labels}
First we gather pillarwise panoptic labels. We then traverse the pillars belonging to the 'thing' classes in the zig-zag order described earlier and if the instance id has not been visited before, i.e., no similar pillars have been visited, we set affinity label 0 to the current pillar. Otherwise we set the label to 1.

\paragraph{Local Clustering for Instance Id Propagation} 
We need to recover the panoptic predictions from the direct outputs of our network, i.e. semantic class predictions and pillar-wise affinity. We assign unique instance ids to pillars with affinity zero and propagate the instance ids to pillars with affinity one (pillars similar to their previous pillars). The algorithm is described in Algorithm \ref{alg}.  We only do local clustering by keeping a memory of instance ids and corresponding pillar locations for the past k columns. As we traverse the pillars, if the pillar is predicted as a `thing' class and has zero affinity, we assign a new instance id corresponding to the semantic class. If the pillar is predicted as a `thing' class and has high affinity to previous pillars, we search the memory for the past k columns and match the current pillar to the instance of the same semantic class and within nearest Manhattan distance in pillar unit. Fig. \ref{fig:clustering} shows an illustration of our binary affinity predictions and the resulting instance id propagation.

\begin{algorithm}
\caption{Local Clustering for Instance Id Propagation}
\KwData{$\mathrm{sem\_pred} \in \{1,2...,K\}^{H\times W},\mathrm{affinity\_pred} \in \{0,1\}^{H\times W}$} 
\KwResult{$\mathrm{ins\_pred}\in \mathbb{Z}^{H\times W}$}
$\mathrm{ins\_pred} \gets 0$ \Comment*[r]{init with ignore class}
$\mathrm{counter}[1,...,L] \gets 0$\Comment*[r]{\#instances for L `thing' classes respectively}

init $\mathbb{M}$ \Comment*[r]{Memory to store instance id and corresponding pillar locations for past k columns}

$\mathrm{offset} \gets 1000$ \Comment*[r]{NuScenes panoptic label format, which sets max \#instance per class to 1000}
\For{$a \gets 0$ to $H-1$}{
  \For{$b \gets 0$ to $W-1$}{
   $s \gets \mathrm{sem\_pred}[a,b]$\;
    \eIf{$s$ is stuff}{
    $\mathrm{ins\_pred}[a,b] \gets s\times \mathrm{offset}$;
  }{\eIf{$\mathrm{affinity\_pred}[a,b]$ is 0}{
      $\mathrm{ins\_pred}[a,b] \gets \mathrm{counter}[s] + 1 + s\times\mathrm{offset}$\;
      $\mathrm{counter}[s] \gets \mathrm{counter}[s]+1$\;
      $update(\mathbb{M}, ins\_pred[a,b], a, b) $\;
    }{$\mathrm{ins\_pred}[a,b] \gets argmin\_distance(\mathbb{M}, s, a, b)$\;
      $update(\mathbb{M}, \mathrm{ins\_pred}[a,b], a, b) $\;
    }
  }
  }
  $drop\_oldest\_column(\mathbb{M}, k)$ \Comment*[r]{keep k columns at maximum}
}
\end{algorithm}\label{alg}

\section{Experiments}
\label{sec:formatting}
\subsection{Dataset and Metrics}
NuScenes\cite{caesar2020nuscenes} is a large scale autonomous driving dataset. It contains 1,000 driving scenes from 4 locations in Singapore and Boston, with 850 scenes for training and validation, and 150 scenes for testing. Points in 40,000 keyframes are manually annotated with 32 fine semantic classes, which can be grouped into 16 coarse semantic classes.

 We use mean intersection over union (mIoU) to evaluate the performance of semantic segmentation. For panoptic segmentation, we use the Panoptic Quality (PQ)\cite{fong2021panoptic}, Semantic Quality (SQ) and Recognition Quality (RQ). 


\begin{table*}
  \centering
  \begin{tabular}{l|ccc|ccc|ccc|c}
    \toprule
    Method & $PQ\uparrow$ &$SQ\uparrow$ & $RQ\uparrow$ & $PQ^{Th}\uparrow$ & $SQ^{Th}\uparrow$ & $RQ^{Th}\uparrow$& $PQ^{St}\uparrow$ & $SQ^{St}\uparrow$ & $RQ^{St}\uparrow$ &seg mIoU$\uparrow$ \\
    \midrule
    sem + box \cite{chen2021polarstream} &77.7&87.2&\textbf{88.7} &\textbf{80.3} &88.9 &\textbf{90} &73.3 &84.2 &86.4 &73\\
    sem + box$\circ$ \cite{chen2021polarstream}&77.1 &87.1 &88 &79.7 &88.8 &89.4 &72.8 &84.2 &85.7 &74.6\\
    \midrule
    sem + centroid \cite{fong2021panoptic}& 76 &87.4 &86.4 &79.2 &\textbf{90.4} &87.6 &70.5 &82.5 &84.6 &70\\
    sem + centroid$\circ$ \cite{fong2021panoptic} &75 &87.3 &85.6 &76.9 &89.6 &85.8 &71.9 &83.5 &85.4 &72.7\\
    \midrule
    sem + affinity (Ours)& 76.7 &87.5 &87.3 &78.5 &89.3 &87.7 &73.6 &84.3 &86.6 &73.6\\
    sem + affinity$\circ$ (Ours) &\textbf{77.9} &\textbf{87.9} &88.2 & 80 &89.7 &89 &\textbf{74.3} &\textbf{84.9} &\textbf{86.9} &\textbf{75.4}\\
    \bottomrule
  \end{tabular}
  \caption{Panoptic Segmentation results on the validation split of nuScenes. $\circ$ denotes polar pillars and otherwise cartesian pillars.}
  \label{tab:comparison}
\end{table*}

\begin{table*}[!h]
  \caption{Class-wise Panoptic Segmentation results on the NuScenes val set. $\circ$ denotes polar pillars and otherwise cartesian pillars.}
  \label{category}
  \centering
  \scalebox{0.8}[0.9]{
  \begin{tabular}{l|cccccccccccccccc|c}
  \hline
  Method & \begin{turn}{90}barrier\end{turn} & \begin{turn}{90}bike\end{turn}  & \begin{turn}{90}bus\end{turn} & \begin{turn}{90}car\end{turn} & \begin{turn}{90}cvehicle\end{turn} & \begin{turn}{90}motorcycle\end{turn} & \begin{turn}{90}pedestrian\end{turn} & \begin{turn}{90}traffic cone\end{turn}  & \begin{turn}{90}trailer\end{turn} & \begin{turn}{90}truck\end{turn}       & \begin{turn}{90}drivable\end{turn} & \begin{turn}{90}other flat\end{turn} & \begin{turn}{90}sidewalk\end{turn}  & \begin{turn}{90}terrain\end{turn} & \begin{turn}{90}manmade\end{turn} & \begin{turn}{90}vegetation\end{turn}& PQ \\
    \midrule
    sem + box \cite{chen2021polarstream} & 72.5	&73.1	&\textbf{81.8} &92.1 &72.3	&87.3	&91.6 &90	&65.4&76.5	&94.9	&56.4	&67.9	&51.3	&\textbf{86.9}	&82.6 &77.7\\
    sem + box$\circ$ \cite{chen2021polarstream} & 72.4	&74.9	&81.4	&\textbf{92.2}	&65.8	&\textbf{88.9}	&\textbf{92.7}	&\textbf{92}	&59.9	&76.7 &95.6	&55.5	&67.4	&51	&86.5	&81.1 &77.1\\
    \midrule
    sem + centroid \cite{fong2021panoptic}& 74.4	&71.6	&80.5	&91.5	&67.2	&82.9	&86.7	&86.7	&\textbf{73.1}	&77.4	&94.4	&54.2	&62.0	&47.7	&85.1	&79.8 &76\\
    sem + centroid$\circ$ \cite{fong2021panoptic}& \textbf{72.6}	&71.7	&77.5	&89.9	&65.8	&81.0	&84.9	&86.3	&68.2	&70.8	&95.1	&55.1	&65.1	&51.4	&85.2	&79.2 &75\\
    \midrule
    sem + affinity (Ours) & 60.3	&73.3	&74.9	&90.1	&\textbf{73.1}	&87.2	&91.3	&90.7	&69.1	&75.1	&95.1	&55.8	&68.5	&52.6	&86.8	&\textbf{82.7} &76.7\\
    sem + affinity$\circ$ (Ours) & 61.3 &\textbf{78.1}	&79.7	&92	&68.8	&87.8	&92.1	&\textbf{92}	&70.4	&\textbf{78}	&\textbf{95.8}	&\textbf{59.1}	&\textbf{68.8}	&\textbf{53.5}	&\textbf{86.9}	&81.5 &\textbf{77.9}\\
    \bottomrule
  \end{tabular}
  }
\end{table*}

\subsection{Baselines}
Our baselines include both proposal-based and proposal-free methods. PolarStream\cite{chen2021polarstream} is the first work for simultaneous lidar object detection and semantic segmentation. The instance ids of points of `thing' classes are associated with the detection box ids whose box centers are the nearest and share the same semantic predictions. Panoptic-PolarNet\cite{fong2021panoptic} is a proposal-free lidar point cloud panoptic segmentation framework that regresses the point centroid of each instance and group the points according to the predicted centroids. For fair comparison, we implement both methods using the same backbone as ours while following their original implementations for everything else.

\subsection{Implementation Details}

Following the same configuration of PolarStream \cite{chen2021polarstream}, we discretize the 3D space within $[distance : 0.3 \sim 50.3m, z: -5 \sim 3m]$ to $[512, 512]$ Polar pillars in NuScenes. For Cartesian pillars, the pillar size is $(0.2, 0.2, 8)m$ within 3D space $[x:-51.2\sim 51.2m, y:-51.2\sim 51.2m, , z:-5 \sim 3m]$, resulting in $[512, 512]$ Cartesian pillars. We implement two sub-task heads, i.e. semantic segmentation and affinity classification with one convolution layer in total, each taking 16 and 2 channels respectively out of 18 channels. We use cross-entropy loss and Lovasz softmax loss \cite{berman2018lovasz} to train semantic segmentation as well as affinity classification. We set both semantic segmentation and affinity classification loss weights to 2. We use adamW \cite{loshchilov2018fixing} optimizer together with one-cycle
policy \cite{smith2018disciplined} with LR max 0.00875, division factor 10, momentum ranges from 0.95 to 0.85, fixed weight decay 0.01 to achieve super convergence. With batch size 56, the model is trained for 20 epochs. We set $k=15$ in instance id propagation.

\subsection{Quantative Results}
Tab. \ref{tab:comparison} shows the comparison between our approach and the baselines on the val split of NuScenes. Our method outperforms the centroid-based proposal-free baseline, Panoptic-PolarNet\cite{fong2021panoptic} (sem + centroid) in both PQ and seg mIoU regardless of the pillar shapes. We also find the offset estimation sub-task in Panoptic-PolarNet diminishes the accuracy on semantic segmentation. It validates the issues of centroid-based approaches as discussed in Sec. \ref{sec:intro} and shows that affinity is a more robust and transferable feature. Our affinity classification head helps improve semantic segmentation and achieves best semantic mIoU among all methods. Affinity classification focuses on structural object information and the discrimination among different objects. Therefore, it is beneficial to semantic segmentation. Our method is slightly better or comparable to the proposal-based baseline PolarStream\cite{chen2021polarstream} (sem+box) in PQ even though PolarStream uses extra 3D box annotations. 

\begin{figure*}[!h]
  \centering
  \begin{subfigure}{0.22\linewidth}
    \includegraphics[scale=0.45]{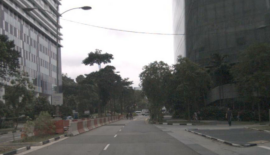} 
    \caption{Original image}
    \label{fig:barrier-a}
  \end{subfigure}
  \hfill
  \begin{subfigure}{0.22\linewidth}
    \includegraphics[scale=0.45]{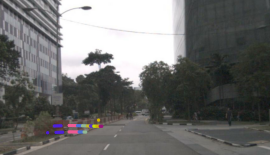} 
    \caption{Ground truth}
    \label{fig:barrier-b}
  \end{subfigure}
   \hfill
  \begin{subfigure}{0.22\linewidth}
    \includegraphics[scale=0.45]{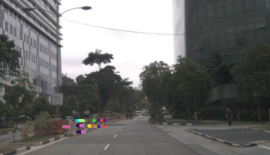}
    \caption{PolarStream\cite{chen2021polarstream} predictions}
    \label{fig:barrier-c}
  \end{subfigure}
  \hfill
  \begin{subfigure}{0.22\linewidth}
    \includegraphics[scale=0.45]{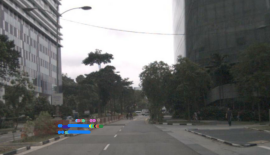} 
    \caption{Our predictions}
    \label{fig:barrier-d}
  \end{subfigure}
  \caption{Visualization of instance ids (color-coded) on barrier class. Our method tends to group connected barriers into fewer instances, resulting in lower PQ but is acceptable in practice. We do all experiments on lidar and project lidar on images here for better visualization.}
  \label{fig:barrier}
\end{figure*}

Tab. \ref{category} shows per-class panoptic quality. Our model lags behind the baselines in the PQ for barriers. We find the reason is that our method groups connected barriers into fewer instances, resulting in more false negatives in evaluation, as shown in Fig. \ref{fig:barrier}. However, this under segmentation is completely acceptable in practice for autonomous driving. Human drivers never try to segment connected barriers.

\begin{table*}[!h]
  \centering
  \begin{tabular}{l|ccc|ccc}
    \toprule
    Method & $PQ\uparrow$ &$SQ\uparrow$ & $RQ\uparrow$ & $PQ^{Th}\uparrow$ & $SQ^{Th}\uparrow$ & $RQ^{Th}\uparrow$ \\
    \midrule
    global clustering &73.3 &85.7 &84.8 &72.7 &86.2 &83.5  \\
    global clustering + iterative &77 &87.7 &87.4 &78.7 &89.4 &87.8 \\
    \midrule
    local clustering  &77.9 &87.9 &88.2 & 80 & 89.7 & 89\\
    \bottomrule
  \end{tabular}
  \caption{Ablation study on choices of clustering.}
  \label{tab:ablation}
\end{table*}

\subsection{Ablation}
To further analyze the influence of clustering for label propagation, we conduct the ablation study on the validation split of NuScenes, as shown in Tab. \ref{tab:ablation}. The results are based on polar pillars. We find global clustering is sub-optimal for our method because our cluster center is the left-most pillar of the instance and grouping a pillar to a cluster center to its right is not reasonable. But if we do global clustering iteratively to update cluster centers by accumulating the points within an instance, the cluster centers are moving towards the centroids, and then global clustering improves by 3.7. Our local clustering matching current pillar to an instance only within past k columns shows the best results.

\subsection{Oracle Tests}
To check what are the upper-bound performances for our model and the baselines, we conduct an oracle test by replacing the predictions of our model with the encoded (pre-processed) and then decoded (post-processed) ground truth and obtain Tab. \ref{tab:upperbound}. We find that provided perfect predictions, all approaches have close PQ for cartesian pillars. For polar pillars, the proposal-free methods show close PQ upper-bound, while having higher upper-bound compared to proposal-based method because it's challenging to do object detection with polar pillars. 
\begin{table}[!htbp]
  \centering
  \begin{tabular}{@{}l|ccc|c@{}}
    \toprule
    Method  & $PQ\uparrow$ & $SQ\uparrow$ & $RQ\uparrow$ &seg mIoU$\uparrow$\\
    \midrule
    sem + box  &92.4 &93.4 &98.9 &92.4 \\
    sem + box$\circ$  & 88.4 &93 &94.7 &95.2\\
    \midrule
    sem + centroid &93.5 &94.3 &99.2 &92.4\\
    sem + centroid$\circ$  &94.7 &95.5 &99.1 &95.2\\
    \midrule
    sem + affinity (ours) &92.6 &94 &98.5 &92.4\\
    sem + affinity$\circ$ (ours)  &94.6 &95.2 &99.4 &95.2\\
    \bottomrule
  \end{tabular}
  \caption{The upper-bound performance of semantic segmentation and panoptic segmentation for the methods. The results are obtained by replacing the predictions of the network with ground truth (encoded and then decoded). PQ: panoptic quality; SQ: semantic quality; RQ: recognition quality; seg mIoU: semantic segmentation mIoU. $\circ$ denotes polar pillars.}
  \label{tab:upperbound}
\end{table}

\begin{table}[!h]
  \centering
  \begin{tabular}{cc|c}
    \toprule
    GT Semantic & GT Affinity & PQ\\
    \midrule
    & &77.9\\
    & \checkmark & 78.9 \\
    \checkmark & & 92.1 \\
    \checkmark &\checkmark & 94.6\\
    \bottomrule
  \end{tabular}
  \caption{Oracle test on the validation split of NuScenes.}
  \label{tab:improvement}
\end{table}

Similar to Panoptic-PolarNet\cite{fong2021panoptic}, we also conducted an oracle test  to investigate the room for improvement in our method, as shown in Tab. \ref{tab:improvement}. We replaced semantic and affinity predictions respectively with ground truth for each experiment and generated the panoptic predictions using the same instance id propagation scheme. It can be seen that our affinity prediction is very close to the ground truth in our test setting and has only -1 difference in PQ compared to given ground truth instance. Conversely, ground truth semantic prediction greatly impacts the results and increases PQ to above 90. This matches the finding in \cite{cheng2020panoptic,fong2021panoptic} that the biggest bottleneck in proposal-free panoptic segmentation is the quality of semantic segmentation predictions.

\section{Conclusion}
In this work, we have shown that lidar panoptic segmentation can be trained as two purely classification sub-tasks, i.e. semantic segmentation and pillar-level affinity prediction, removing the need for object detection heads and their associated post-processing operations like NMS. We also show that pillar-level affinities are a more robust feature to learn for the panoptic segmentation task as compared to learning instance centroids as the former is synergistic to the semantic segmentation task. We finally propose a simple and efficient local clustering algorithm to propagate instance ids by merging semantic segmentation and affinity predictions. Our model outperforms previous proposal-free methods and completely closes the gap between proposal-based and proposal-free methods.


{\small
\bibliographystyle{ieee_fullname}
\bibliography{egbib}

\begin{thebibliography}{10}\itemsep=-1pt

\bibitem{ahn2018learning}
Jiwoon Ahn and Suha Kwak.
\newblock Learning pixel-level semantic affinity with image-level supervision
  for weakly supervised semantic segmentation.
\newblock In {\em Proceedings of the IEEE conference on computer vision and
  pattern recognition}, pages 4981--4990, 2018.

\bibitem{berman2018lovasz}
Maxim Berman, Amal~Rannen Triki, and Matthew~B Blaschko.
\newblock The lov{\'a}sz-softmax loss: A tractable surrogate for the
  optimization of the intersection-over-union measure in neural networks.
\newblock In {\em Proceedings of the IEEE conference on computer vision and
  pattern recognition}, pages 4413--4421, 2018.

\bibitem{bertasius2017convolutional}
Gedas Bertasius, Lorenzo Torresani, Stella~X Yu, and Jianbo Shi.
\newblock Convolutional random walk networks for semantic image segmentation.
\newblock In {\em Proceedings of the IEEE Conference on Computer Vision and
  Pattern Recognition}, pages 858--866, 2017.

\bibitem{caesar2020nuscenes}
Holger Caesar, Varun Bankiti, Alex~H Lang, Sourabh Vora, Venice~Erin Liong,
  Qiang Xu, Anush Krishnan, Yu Pan, Giancarlo Baldan, and Oscar Beijbom.
\newblock nuscenes: A multimodal dataset for autonomous driving.
\newblock In {\em Proceedings of the IEEE/CVF conference on computer vision and
  pattern recognition}, pages 11621--11631, 2020.

\bibitem{chen2021polarstream}
Qi Chen, Sourabh Vora, and Oscar Beijbom.
\newblock Polarstream: Streaming object detection and segmentation with polar
  pillars.
\newblock {\em Advances in Neural Information Processing Systems}, 34, 2021.

\bibitem{cheng2020panoptic}
Bowen Cheng, Maxwell~D Collins, Yukun Zhu, Ting Liu, Thomas~S Huang, Hartwig
  Adam, and Liang-Chieh Chen.
\newblock Panoptic-deeplab: A simple, strong, and fast baseline for bottom-up
  panoptic segmentation.
\newblock In {\em Proceedings of the IEEE/CVF conference on computer vision and
  pattern recognition}, pages 12475--12485, 2020.

\bibitem{cheng2017locality}
Yanhua Cheng, Rui Cai, Zhiwei Li, Xin Zhao, and Kaiqi Huang.
\newblock Locality-sensitive deconvolution networks with gated fusion for rgb-d
  indoor semantic segmentation.
\newblock In {\em Proceedings of the IEEE conference on computer vision and
  pattern recognition}, pages 3029--3037, 2017.

\bibitem{fong2021panoptic}
Whye~Kit Fong, Rohit Mohan, Juana~Valeria Hurtado, Lubing Zhou, Holger Caesar,
  Oscar Beijbom, and Abhinav Valada.
\newblock Panoptic nuscenes: A large-scale benchmark for lidar panoptic
  segmentation and tracking.
\newblock {\em arXiv preprint arXiv:2109.03805}, 2021.

\bibitem{he2017mask}
Kaiming He, Georgia Gkioxari, Piotr Doll{\'a}r, and Ross Girshick.
\newblock Mask r-cnn.
\newblock In {\em Proceedings of the IEEE international conference on computer
  vision}, pages 2961--2969, 2017.

\bibitem{hu2021istr}
Jie Hu, Liujuan Cao, Yao Lu, ShengChuan Zhang, Yan Wang, Ke Li, Feiyue Huang,
  Ling Shao, and Rongrong Ji.
\newblock Istr: End-to-end instance segmentation with transformers.
\newblock {\em arXiv preprint arXiv:2105.00637}, 2021.

\bibitem{Kirillov_2019_CVPR}
Alexander Kirillov, Ross Girshick, Kaiming He, and Piotr Dollar.
\newblock Panoptic feature pyramid networks.
\newblock In {\em Proceedings of the IEEE/CVF Conference on Computer Vision and
  Pattern Recognition (CVPR)}, June 2019.

\bibitem{lang2019pointpillars}
Alex~H Lang, Sourabh Vora, Holger Caesar, Lubing Zhou, Jiong Yang, and Oscar
  Beijbom.
\newblock Pointpillars: Fast encoders for object detection from point clouds.
\newblock In {\em Proceedings of the IEEE/CVF Conference on Computer Vision and
  Pattern Recognition}, pages 12697--12705, 2019.

\bibitem{lin2017feature}
Tsung-Yi Lin, Piotr Doll{\'a}r, Ross Girshick, Kaiming He, Bharath Hariharan,
  and Serge Belongie.
\newblock Feature pyramid networks for object detection.
\newblock In {\em Proceedings of the IEEE conference on computer vision and
  pattern recognition}, pages 2117--2125, 2017.

\bibitem{loshchilov2018fixing}
Ilya Loshchilov and Frank Hutter.
\newblock Fixing weight decay regularization in adam.
\newblock 2018.

\bibitem{ronneberger2015u}
Olaf Ronneberger, Philipp Fischer, and Thomas Brox.
\newblock U-net: Convolutional networks for biomedical image segmentation.
\newblock In {\em International Conference on Medical image computing and
  computer-assisted intervention}, pages 234--241. Springer, 2015.

\bibitem{sirohi2021efficientlps}
Kshitij Sirohi, Rohit Mohan, Daniel B{\"u}scher, Wolfram Burgard, and Abhinav
  Valada.
\newblock Efficientlps: Efficient lidar panoptic segmentation.
\newblock {\em IEEE Transactions on Robotics}, 2021.

\bibitem{smith2018disciplined}
Leslie~N Smith.
\newblock A disciplined approach to neural network hyper-parameters: Part
  1--learning rate, batch size, momentum, and weight decay.
\newblock {\em arXiv preprint arXiv:1803.09820}, 2018.

\bibitem{wang2020axial}
Huiyu Wang, Yukun Zhu, Bradley Green, Hartwig Adam, Alan Yuille, and
  Liang-Chieh Chen.
\newblock Axial-deeplab: Stand-alone axial-attention for panoptic segmentation.
\newblock In {\em European Conference on Computer Vision}, pages 108--126.
  Springer, 2020.

\bibitem{xiong2019upsnet}
Yuwen Xiong, Renjie Liao, Hengshuang Zhao, Rui Hu, Min Bai, Ersin Yumer, and
  Raquel Urtasun.
\newblock Upsnet: A unified panoptic segmentation network.
\newblock In {\em Proceedings of the IEEE/CVF Conference on Computer Vision and
  Pattern Recognition}, pages 8818--8826, 2019.

\end{thebibliography}
}

\end{document}